\documentclass[10pt,twocolumn,letterpaper]{article}

\usepackage{wacv}
\usepackage{times}
\usepackage{epsfig}
\usepackage{graphicx}
\usepackage{amsmath}
\usepackage{amssymb}
\usepackage{booktabs}

\usepackage{xcolor}
\usepackage{epsfig}
\usepackage{graphicx}
\usepackage{amsmath}
\usepackage{enumitem}

\usepackage[labelfont=bf]{caption}

\usepackage{physics}
\usepackage{graphicx}
\usepackage{float}
\usepackage{amsmath}
\usepackage{subfigure}
\usepackage{amsmath}
\usepackage{longtable}
\usepackage{booktabs}
\usepackage{float}
\usepackage{threeparttable}
\usepackage{xcolor}
\usepackage{diagbox}
\usepackage{hyperref}

\usepackage{amsfonts}
\usepackage{amsmath}
\usepackage{amssymb}

\usepackage{multirow}
\usepackage{multicol}
\usepackage{tabularx}
\usepackage{dsfont}
\usepackage{url}

\usepackage{amsthm}
\usepackage[export]{adjustbox}
\usepackage{comment}
\usepackage{cite}
\usepackage{hyperref}

\usepackage[belowskip=1pt,aboveskip=0pt]{caption}

\usepackage[utf8]{inputenc} 
\usepackage[T1]{fontenc}    
\usepackage{url}            
\usepackage{booktabs}       
\usepackage{amsfonts}       
\usepackage{nicefrac}       
\usepackage{microtype}      

\definecolor{dg}{rgb}{0,0.694,0.298}
\definecolor{purple}{rgb}{0.4,0.176,0.569}

\usepackage{pifont}
\usepackage{xspace}
\makeatletter
\DeclareRobustCommand\onedot{\futurelet\@let@token\@onedot}
\def\@onedot{\ifx\@let@token.\else.\null\fi\xspace}

\makeatother

\usepackage[accsupp]{axessibility} 

\def\wacvPaperID{837} 

\wacvfinalcopy 

\begin{document}

\title{Domain Adaptive Object Detection for Autonomous Driving under Foggy Weather}

\author{Jinlong~Li$^1$,~Runsheng~Xu$^2$,~Jin~Ma$^1$,~Qin~Zou$^3$,~Jiaqi~Ma$^2$,~Hongkai~Yu$^{1}$\thanks{  Corresponding author}
\\
$^1$~Cleveland State University, $^2$~University of California, Los Angeles, $^3$~Wuhan University\\
{\tt\small j.li56@vikes.csuohio.edu, h.yu19@csuohio.edu}
}


\maketitle
\thispagestyle{empty}

\begin{abstract}
Most object detection methods for autonomous driving usually assume a consistent feature distribution between training and testing data, which is not always the case when weathers differ significantly. The object detection model trained under clear weather might be not effective enough on the foggy weather  because of the domain gap. This paper proposes a novel domain adaptive object detection framework  for autonomous driving under foggy weather. Our method leverages both image-level and object-level adaptation to diminish the domain discrepancy in image style and object appearance. To further enhance the model's  capabilities under challenging samples, we also come up with a new adversarial gradient reversal layer to perform adversarial mining for the hard examples together with domain adaptation. Moreover, we propose to generate an auxiliary domain by data augmentation to enforce a new domain-level metric regularization. Experimental results on public benchmarks show the effectiveness and accuracy of the proposed method. The code is available at \url{https://github.com/jinlong17/DA-Detect}.
\end{abstract}
    
\vspace{-1em}
\section{Introduction}
Autonomous driving has wide applications for intelligent transportation systems, such as improving the efficiency in the automatic 24/7 working manner, reducing the labor costs, enhancing the comfortableness of customers, and so on~\cite{jo2014development,xu2021opencda}. With the computer vision and artificial intelligence techniques, object detection plays a critical role in autonomous driving to understand the surrounding driving scenarios~\cite{xu2022opencood,zhao2020fusion}. In some cases, the autonomous vehicle might work in the complex residential and industry areas. The diverse weather conditions might make the object detection in these environments more difficult. For example, the usages of heating, gas, coal, and vehicle emissions in residential and industry areas  might be possible to generate more frequent foggy or hazy weather, leading to a significant challenge to the object detection system installed on the autonomous   vehicle.

\begin{figure*} 
    \begin{centering}
        \includegraphics[width=1\textwidth]{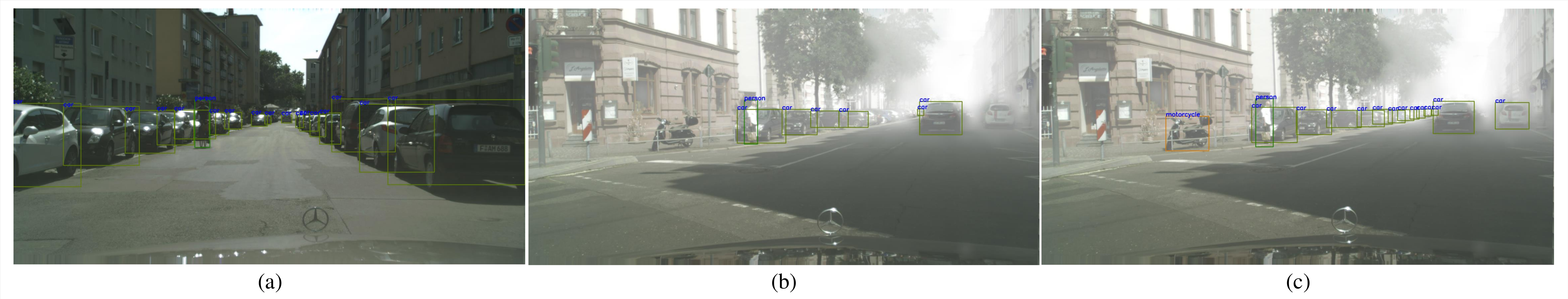}
    \par\end{centering}
    \caption{Illustration of the domain adaptive object detection for autonomous driving: (a) Faster R-CNN~\cite{ren2015faster} detection under clear weather, (b) Faster R-CNN detection under foggy weather without domain adaptation, (c) Faster R-CNN detection under foggy weather with the proposed domain adaptation.}
    \label{fig:introduction}
\end{figure*}

Many deep learning models such as Faster R-CNN~\cite{ren2015faster}, YOLO~\cite{redmon2016you} have demonstrated great success in autonomous driving. However, most of these well-known methods assume that the feature distributions of training and testing data are homogeneous. Such an assumption may fail when taking the real-world diverse weather conditions into account~\cite{sakaridis2018semantic}. For example, as shown in Fig.~\ref{fig:introduction}, the Faster R-CNN model trained on the clear-weather data (source domain) is capable of detecting objects accurately under good weather, but its performance drops significantly when it comes to the foggy weather (target domain). This degradation is caused by the feature domain gap between divergent weather conditions, as the model is unfamiliar with the feature distribution on the target domain, while the detection performance could be improved under the foggy weather with domain adaptation.

Domain adaptation, as a technique of transfer learning, is to reduce the domain shift between various weathers. This paper proposes a novel domain adaptation framework to achieve robust object detection performance in autonomous driving under foggy weather. As manually annotating images under adverse weathers is usually time-consuming, our design follows an unsupervised fashion same as that  in~\cite{chen2018domain,song2020multi,li2021domain}, where clear-weather images (source domain) are  well labeled and foggy weather images (target domain) have no annotations. Inspired by~\cite{chen2018domain,guan2021uncertainty}, our method leverages both image-level and object-level adaptation to diminish the domain discrepancy in image style and object appearance jointly, which is realized by involving image-level and object-level domain classifiers to enable our convolutional  neural networks generating domain-invariant latent feature representations. Specifically, the domain classifiers aim to maximize the probability of distinguishing the features produced by different domains, whereas the detection model expects to generate the domain-invariant features to confuse the classifiers. 

This paper also addresses two critical insights that are ignored by previous domain adaptation methods~\cite{chen2018domain,li2021domain,fu2021let,zheng2020cross,guan2021uncertainty}: 1) Different training samples might have different challenging levels to be fully harnessed during the transfer learning, while existing works usually omit such diversity; 2) Previous domain adaptation methods only consider the source domain and target domain for transfer learning, while the domain-level feature metric distance to the third related domain might be neglected. However, embedding the mining for hard examples and involving an extra related domain might potentially further enhance the model's robust learning capabilities, which has not been carefully explored before. To emphasize these two insights,  we propose a new Adversarial Gradient Reversal Layer (AdvGRL) and generate an auxiliary domain by data augmentation. The AdvGRL performs adversarial mining for the hard examples to enhance the model learning  on the challenging scenarios, and the auxiliary domain enforces a new domain-level metric regularization during the transfer learning. Experimental results on the public benchmarks  Cityscapes~\cite{cordts2016cityscapes} and Foggy Cityscapes~\cite{sakaridis2018semantic} show the effectiveness of each proposed component and the superior object detection performance over the baseline and comparison methods. Overall, the contributions of this paper are summarized as follows:

\begin{itemize}
    \item We propose a novel deep transfer learning based domain adaptive object detection framework for autonomous driving under foggy weather, including the image-level and object-level adaptations, which is trained with labeled clear-weather data and unlabeled foggy-weather data to enhance the generalization ability of the deep learning based object detection model.   
    
    \item We propose a new Adversarial Gradient Reversal Layer (AdvGRL) to perform adversarial mining for the hard examples together with the domain adaptation to further enhance the model's transfer learning capabilities under challenging samples.
    
    \item We propose a new domain-level metric regularization during the transfer learning. By generating an auxiliary domain with data augmentation, the domain-level metric constraint between source domain, auxiliary domain, and target domain is ensured as regularization during the transfer learning.    
     
\end{itemize}


\section{Related Work}~\label{Sec:Related Work}
\vspace{-2em}
\subsection{Object detection for autonomous driving}
Recent advancement in deep learning has brought outstanding progress in autonomous driving~\cite{paz2020probabilistic,christensen2021autonomous,li2020sus,xu2021holistic}, and object detection has been one of the most active topic under this field~\cite{shan2019pixel,zhao2020fusion,feng2021review,tu2022maxvit}. Regarding the network architecture, current object detection algorithms can be roughly split into two categories: two-stage methods and single-stage methods. Two-stage object detection algorithms typically compose of two processes: 1) region proposal, 2) object classification and localization refinement.  R-CNN~\cite{girshick2014rich} is the first work for this kind of methods, it applies selective search for regional proposals and independent CNNs for each object prediction. Fast R-CNN~\cite{girshick2015fast} improves R-CNN by obtaining object features from the shared feature map learned by one CNN. Faster R-CNN~\cite{ren2015faster} further enhances the framework by proposing Region Proposal Network (RPN) to replace the selective search stage. Single-stage object detection algorithms predict object bounding boxes and classes simultaneously in one same stage. These methods usually leverage pre-defined anchors to classify objects and regress bounding boxes, they are less time-consuming but less accurate compared to two-stage algorithms. Milestones for this category include SSD-series~\cite{liu2016ssd}, YOLO-series~\cite{redmon2016you} and RetinaNet~\cite{lin2017focal}. Despite their success in clear-weather visual scenes, these object detection methods might not be employed in autonomous driving directly due to the complex real-world weather conditions. 

\subsection{Object detection for autonomous driving under different weather}

In order to address the diverse weather conditions encountered in autonomous driving, many datasets have been generated~\cite{sakaridis2018semantic,michaelis2019benchmarking,pang2020tju,pham20203d} and many methods have been proposed~\cite{huang2020dsnet,hahner2021fog,qian2021robust,bijelic2020seeing,sindagi2020prior,tu2022maxim,he2019multi} in recent years. For example, Foggy Cityscape~\cite{sakaridis2018semantic} is a synthetic dataset that applies fog simulation to Cityscape for scene understanding in foggy weather. TJU-DHD~\cite{pang2020tju} is a diverse dataset for object detection in real-world scenarios which contains variances in terms of illumination, scene, weather and season. In this paper, we focus on the object detection problem in foggy weather. Huang et al.~\cite{huang2020dsnet} propose a DSNet (Dual-Subnet Network) that involves a detection subnet and a restoration subnet. This network can be trained with multi-task learning by combining visibility enhancement task and object detection task, thus outperforms pure object detectors. Hahner et al.~\cite{hahner2021fog} develop a fog simulation approach to enhance existing real lidar dataset, and show this approach can be leveraged to improve current object detection methods in foggy weather. Qian et al.~\cite{qian2021robust} propose a MVDNet (Multimodal Vehicle Detection Network) that takes advantage of lidar and radar signals to obtain proposals. Then the region-wise features from these two sensors are fused together to get final detection results. Bijelic et al.~\cite{bijelic2020seeing} develop a network that takes the data from four sensors as input: lidar, RGB camera, gated camera, and radar. This architecture uses entropy-steered adaptive deep fusion to get fused feature maps for prediction. These methods typically rely on input data from other sensors rather than RGB camera itself, which is not the general case for many autonomous driving cars. Thus we aim to develop an object detection architecture that only takes RGB camera data as input in this work. 

\subsection{Domain adaptation for object detection}
Domain adaptation reduces the discrepancy between different domains, thus allows the model trained on source domain to be applicable on unlabeled target domain. Previous domain adaptation works mainly focus on the task of image classification~\cite{tzeng2017adversarial,wang2018deep,you2019universal,xiao2021dynamic}, while more and more methods have been proposed to solve domain adaptation for object detection in recent years~\cite{chen2018domain,kim2019diversify,saito2019strong,zhao2020adaptive,xu2021spg,yang2021st3d,zhang2021rpn,xu2020cross,guan2021uncertainty}. Domain adaptive detectors could be obtained if the features from different domains are aligned~\cite{chen2018domain,he2019multi,saito2019strong,xu2020cross,guan2021uncertainty, xu2022bridging}. From this perspective, Chen et al.~\cite{chen2018domain} introduce a Domain Adaptive Faster R-CNN framework to reduce domain gap from image level and instance level, and the image-and-instance consistency is subsequently employed to improve cross-domain robustness. He et al.~\cite{he2019multi} propose a MAF (multi-adversarial Faster R-CNN) framework to minimize the domain distribution disparity by aligning domain features and proposal features hierarchically. On the other hand, some works try to solve domain adaptation through image style transfer methods~\cite{shan2019pixel,kim2019diversify,hsu2020progressive}. Shan et al.~\cite{shan2019pixel} first convert images from source domain to target domain with image translation module, then train the object detector with adversarial training on target domain. Hsu et al.~\cite{hsu2020progressive} choose to translate images progressively, and add a weighted task loss during adversarial training stage for tackling the problem of image quality difference.  Many previous methods~\cite{zhou2022multi, li2022sigma, rezaeianaran2021seeking,chen2022learning} design complex architectures. \cite{zhou2022multi} used multi-scale backbone Feature Pyramid Networks and considered pixel-level and category-level adaptation. \cite{li2022sigma} used the complex Graph Convolution Network and  graph matching algorithms.  \cite{rezaeianaran2021seeking} used the similarity-based clustering and grouping. 
\cite{chen2022learning} uses the uncertainty-guided self-training mechanism (Probabilistic Teacher and Focal Loss) to capture the uncertainty of unlabeled target data from a gradually evolving teacher and guides student learning. Differently, our method does not bring extra learnable parameters to original Faster R-CNN model because our AdvGRL is based on adversarial training (gradient reversal) and Domain-level Metric  Regularization is based on triplet loss. Previous domain adaptation methods usually treat training samples at the same challenging level, while we employ advGRL  for adversarial hard example mining  to improve transfer learning. Moreover, we generate an auxiliary domain and apply domain-level metric regularization to avoid overfitting. 

\section{Proposed Method}~\label{Sec:Method}
In this section, we will first introduce the overall network architecture, then describe the image-level and object-level adaptation method, and finally, reveal the details of AdvGRL and domain-level metric regularization.

\subsection{Network Architecture}

\begin{figure*}[htb]
	\begin{minipage}[b]{1\textwidth}
		\centering
		\includegraphics[width=0.9\textwidth]{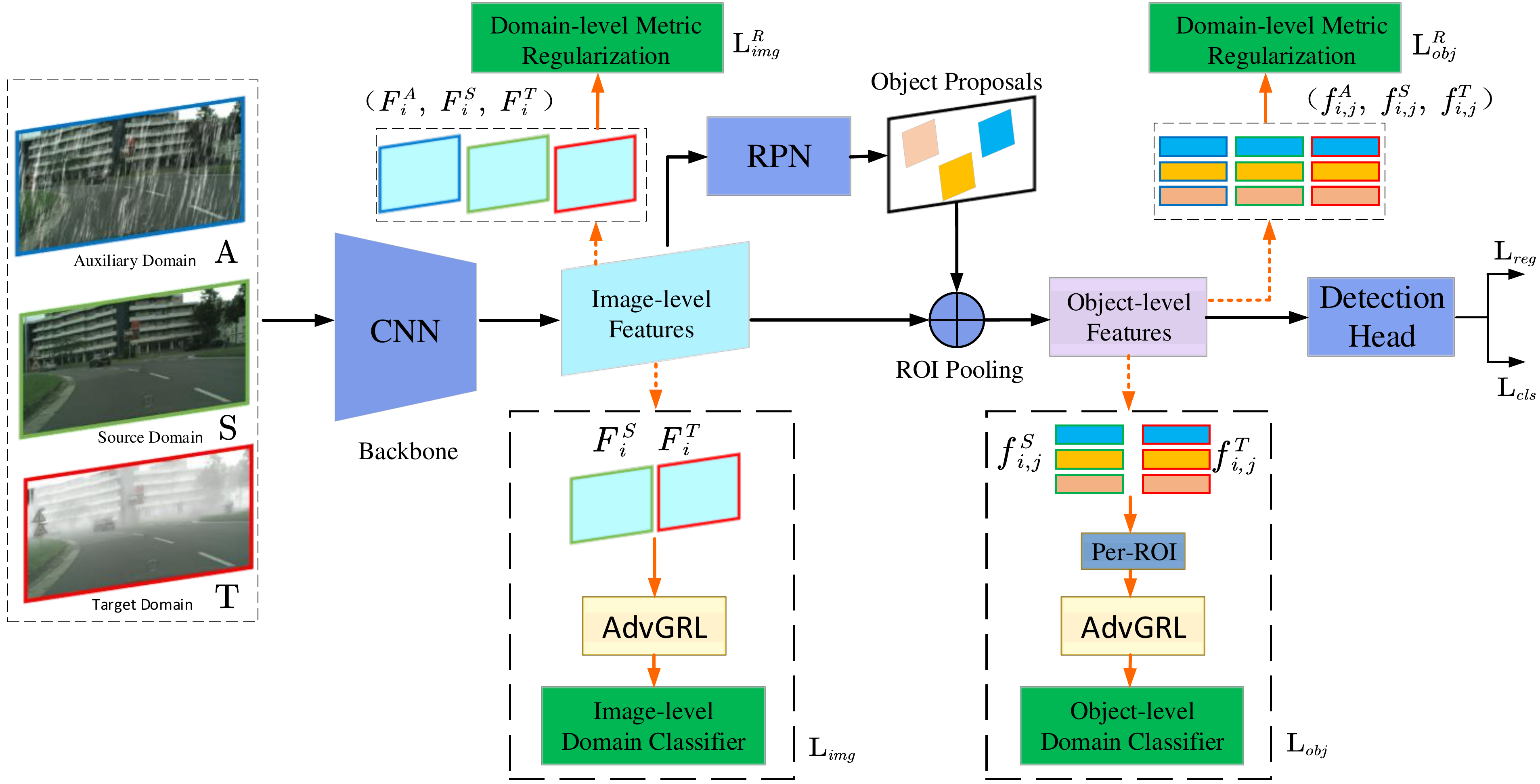}
	\end{minipage}
	\caption{The architecture of proposed domain adaptive object detection for autonomous driving under foggy weather. Based on the traditional Faster R-CNN architecture~\cite{ren2015faster}, the image-level and object-level domain adaptations with adversarial gradient reversal layer (AdvGRL) and  domain-level metric regularization are designed in the proposed  framework. This figure is best viewed in color.} 
	\label{fig:da_faster_rcnn} 
\end{figure*}

As illustrated in Fig.~\ref{fig:da_faster_rcnn}, our proposed model adopts the pipeline in Faster R-CNN for object detection. The Convolutional Neural Network (CNN) backbone extracts the image-level features from the RGB images and send them to Region Proposal Network (RPN) to generate object proposals. Afterwards, the ROI pooling accepts both image-level features and object proposals as the input to retrieve the object-level features. Eventually, a detection head is applied on the object-level features to produce the final predictions. Based on the Faster R-CNN framework, we integrate two more components: image-level domain adaptation module, and object-level domain adaptation module. For both modules, we deploy a new Adversarial Gradient Reversal Layer (AdvGRL) together with the domain classifier to extract domain-invariant features and perform adversarial hard example mining. Moreover, we involve an auxiliary domain to impose a new domain-level metric regularization to enforce the feature metric distance between different domains. All three domains, \textit{i.e.}, source, target, and auxiliary domains, will be employed simultaneously during the training.

\subsection{Image-level Adaptation}

The image-level domain representation is obtained from the backbone feature extraction and contains rich global information such as style, scale and illumination, which can potentially pose significant impacts on the detection task~\cite{chen2018domain}. Therefore, a domain classifier is introduced to classify the domains of the upcoming image-level features to enhance the image-level global alignment.
The domain classifier is just a simple CNN with two convolutional layers and it will output a prediction to identify the feature domain. We use the binary cross entropy loss for the domain classifier as follows:

\begin{equation}\label{equ:img_loss}
 L_{img} = - \sum_{i=1}^{N}[G_{i} {\rm log} P_{i} + (1-G_{i}) {\rm log}(1-P_{i})], 
\end{equation}
where $i\in\{1,..., N\}$ represents the $N$ training images, $G_i\in\{1, 0\}$ is the ground truth of the domain label in the $i$-th training image ($1$ and $0$ stand for source and target domains respectively), and $P_i$ is the prediction of the domain classifier.

\subsection{Object-level Adaptation} 
Besides the image-level global difference in different domains, the objects in different domains might be also dissimilar in the appearance, size, color, \textit{etc}. In this paper, we define each region proposal after the ROI Pooling layer in Faster R-CNN as a potential object. Similar with image-level adaptation module, after retrieving the object-level domain representation by ROI pooling, we implement a object-level domain classifier to identify the feature derivation from local information. A well-trained object-level classifier, a neural network with 3 fully-connected layers, will help align the object-level feature distribution. We also use the binary cross entropy loss for this domain classifier: 

\begin{equation}\label{equ:obj_loss}
 L_{obj} = - \sum_{i=1}^{N} \sum_{j=1}^{M}[G_{i,j} {\rm log} P_{i,j} + (1-G_{i,j}) {\rm log}(1-P_{i,j})], 
\end{equation}
where $j\in\{1,...,M\}$ is the $j$-th detected object (region proposal) in the $i$-th image, $P_{i,j}$ is the prediction of the object-level domain classifier for the $j$-th region proposal in the $i$-th image, and $G_{i,j}$ is the corresponding binary ground-truth label for source and target domains respectively. 



\subsection{Adversarial Gradient Reversal Layer}\label{Subsec:AdvGRL}

\begin{figure}[htb]
	\centering
	\includegraphics[width=0.7\columnwidth]{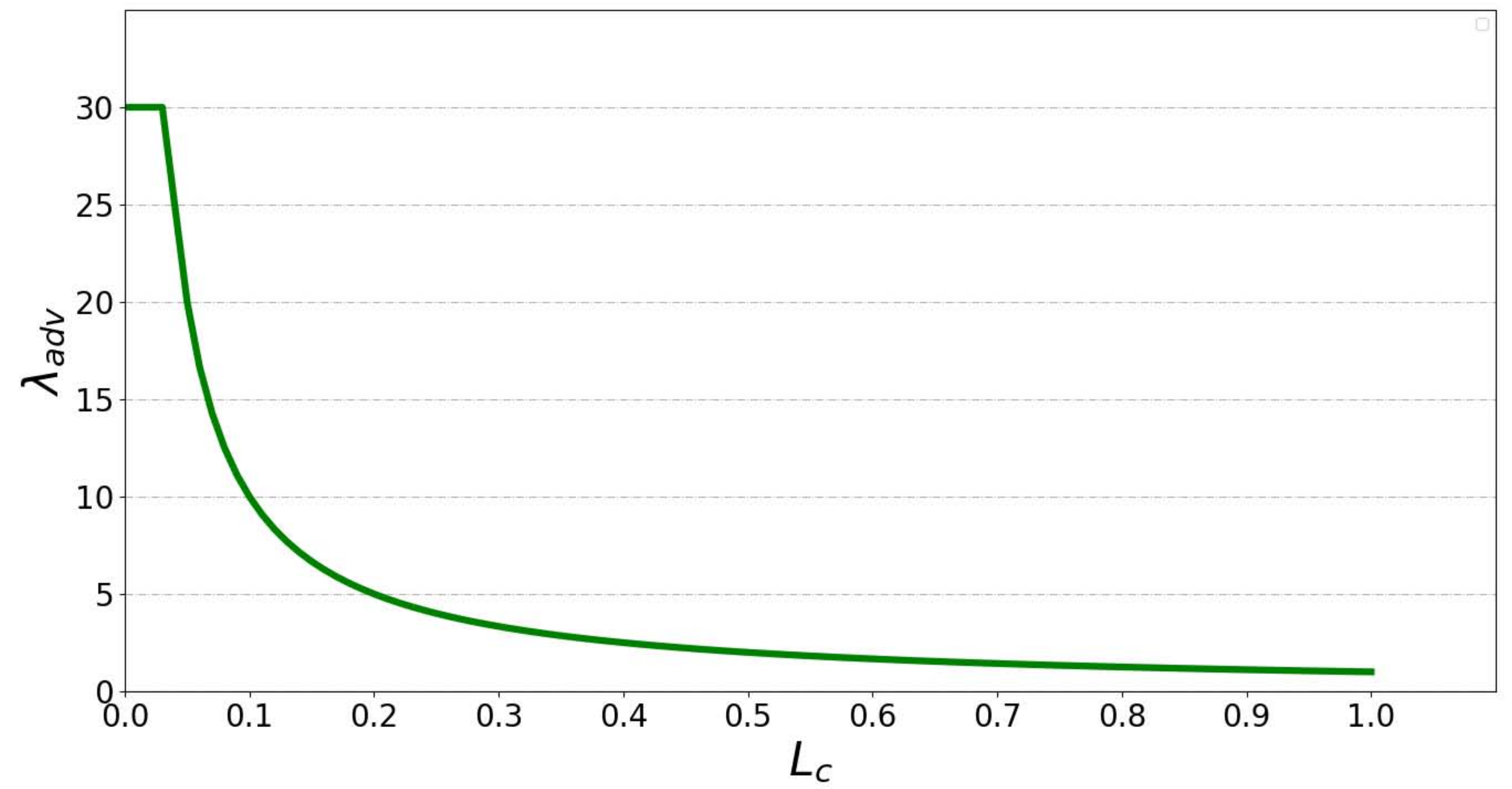}
	\caption{Illustration of the adversarial mining for hard training examples by the proposed AdvGRL. In this example, we set $\lambda_{0}=1$, $\beta=30$. 
	Harder training examples with lower domain classifier loss $L_c$ will have larger response.} 
	\label{fig:grl_loss} 
\end{figure}

\begin{figure}[htbp]
	\centering
	\includegraphics[width=1\columnwidth]{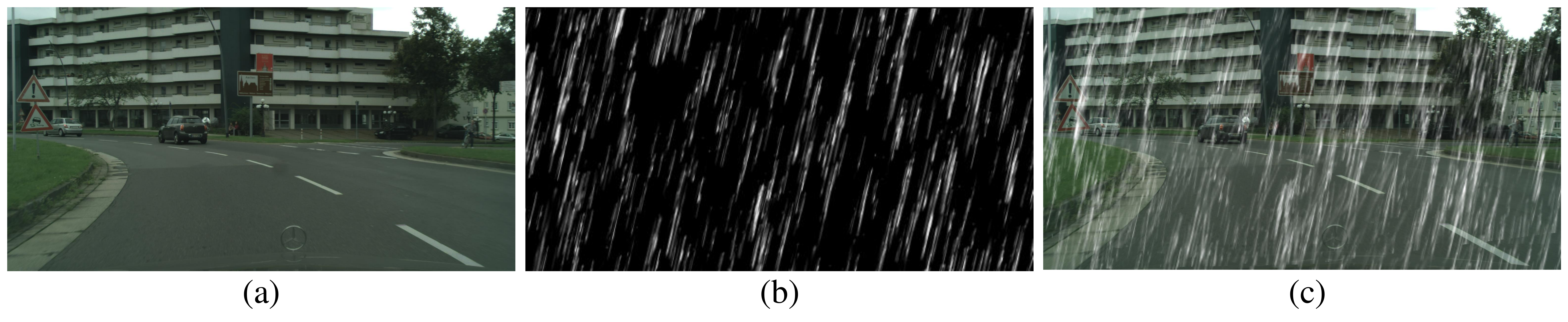}
	\caption{The example of generating auxiliary domain using Cityscapes dataset: (a) an original image, (b) a rain map by RainMix~\cite{guo2021efficientderain}, (c) a synthetic rainy Cityscapes image for the auxiliary domain.} 
	\label{fig:rain_domain} 
\end{figure}

In this section, we first review the original Gradient Reversal Layer (GRL)~\cite{ganin2015unsupervised}, then we make a detailed description of the proposed Adversarial Gradient Reversal Layer (AdvGRL) for our domain adaptive object detection framework.  

The original GRL is used for unsupervised domain adaptation of the image classification task~\cite{ganin2015unsupervised}. Specifically, it leaves the input unchanged during forward propagation and reverses the gradient by multiplying it by a negative scalar when back-propagating to the base network ahead during training. A domain classifier is trained to maximize the probability of identifying the domain while the base network ahead is optimized to confuse the domain classifier. In this way, the domain-invariant features are obtained to realize the domain adaptation. The forward propagation of GRL is defined as:  

\begin{equation}\label{equ:grl_1}
  R_{\lambda}({\mathbf v}) = {\mathbf v},
\end{equation}
where ${\mathbf v}$ is an input feature vector, and $R_{\lambda}$ denotes the forwarding function that GRL performs, and the back-propagation of GRL is defined as:

\begin{equation}\label{equ:grl_2}
  \frac{dR_{\lambda}}{d{\mathbf v}} = -\lambda \mathbf{I}, 
\end{equation}
where $\mathbf{I}$ is an identity matrix and $-\lambda$ is a negative scalar.

The original GRL sets either a constant or a changing $-\lambda$ based on the training iterations~\cite{ganin2015unsupervised}. However, this setting ignores an insight that different training samples might have different challenging levels during the transfer learning.  Therefore, this paper proposes a novel AdvGRL to perform adversarial mining for the hard examples together with the domain adaptation to further enhance the model's transfer learning capabilities under   challenging examples. This can be done by simply replacing $\lambda$ by a new $\lambda_{adv}$ in Eq.~(\ref{equ:grl_2})  of GRL, which forms the proposed AdvGRL. Particularly, $\lambda_{adv}$ is calculated as:


\begin{equation}\label{equ:ad_grl}
  \lambda_{adv} = \left \{
    \begin{aligned}
        & {\rm min} (\frac{\lambda_{0}}{L_{c}}, \beta), \qquad  & L_{c} < \alpha \\
        &\lambda_{0},   \qquad   & {\rm otherwise}, 
    \end{aligned}
    \right.
\end{equation}
where $L_{c}$ is the loss of the domain classifier,  $\alpha$ is a hardness threshold to judge whether the training sample is challenging, $\beta$ is the overflow threshold to avoid generating excessive gradients in the back-propagation, and $\lambda_{0}=1$ is set as a fixed parameter in our experiment. In other words, if the domain classifier's loss $L_{c}$ is smaller, the domain of the training sample can be more easily identified, whose feature is not the desired domain-invariant feature, so this kind of training sample is a harder example for domain adaptation. The  relation of $\lambda_{adv}$ and $L_{c}$ is shown in Fig.~\ref{fig:grl_loss}. 

On summary, the proposed AdvGRL has two effects: 1) AdvGRL could use negative gradients during back-propagation to confuse the domain classifier so as to generate domain-invariant features; 2) AdvGRL could perform adversarial mining for the hard examples to further enhance the model generalization 
under challenging examples. The proposed AdvGRL is applied to both image-level and object-level domain adaptation in our domain adaptive object detection framework, as shown in Fig.~\ref{fig:da_faster_rcnn}.

\subsection{Domain-level Metric Regularization}

Previous existing domain adaptation methods mainly focus on the transfer learning from source domain $S$ to target domain $T$, which neglects the potential benefits of the third related domain can bring. To address this and thus additionally involve the feature metric constraint between different domains, we introduce an auxiliary domain for a domain-level metric regularization during the transfer learning.

Based on the source domain $S$, we can apply some advanced data augmentation methods to generate an auxiliary domain $A$. For the autonomous driving scenario, the training data in different weather conditions can be synthesized from the clear-weather data, then the three input images of our architecture (as shown in Fig.~\ref{fig:da_faster_rcnn}) could be aligned images. For example, we generate an auxiliary domain with the advanced data augmentation method  RainMix~\cite{guo2021efficientderain,hendrycks2020augmix}. 
Specifically, we randomly sample a rain map from the public dataset of real rain streaks~\cite{garg2006photorealistic}, then perform  random transformations using the RainMix technique on the rain map, where these random transformations (\textit{i.e.}, rotate, zoom, translate, shear) are sampled and combined. Finally, these transformed rain maps can be blended with the source domain images, which can simulate the diverse rain patterns in the real world. The example of generating auxiliary domain is shown in Fig.~\ref{fig:rain_domain}. Different with other methods including data augmentation to the source/target domain, by generating an  auxiliary domain with data augmentation, the domain-level metric constraint between source, auxiliary, and target domains is ensured.

Let us define the $i$-th training image's global image-level features of $S$, $A$ and $T$ as $F_{i}^S$, $F_{i}^{A}$, and $F_{i}^{T}$ respectively. We expect to ensure the feature metric distance between the $F_{i}^S$ and $F_{i}^{T}$ closer than the feature metric distance between $F_{i}^S$ and $F_{i}^{A}$ after reducing the domain gap between $S$ and $T$, which is defined as:

\begin{equation}\label{equ:reg_disance}
  d(F_{i}^{S}, F_{i}^{T}) < d(F_{i}^{S}, F_{i}^{A}),
\end{equation}
where $d(,)$ denotes the metric distance of the corresponding features. This constraint can be implemented by a triplet structure, where the $F_{i}^S$, $F_{i}^{T}$, $F_{i}^{A}$ can be treated as anchor, positive and negative in the triplet structure. Therefore, as the domain-level metric regularization on image features, the above image-level constraint in Eq.~(\ref{equ:reg_disance}) is equivalent to minimize the following image-level triplet loss:    

\begin{equation}\label{equ:reg_loss_img}
  L^{R}_{img} = {\rm max}(d(F_{i}^{S}, F_{i}^{T} ) - d(F_{i}^{S}, F_{i}^{A}) + \delta , 0), 
\end{equation}
where the parameter $\delta$ is used as a margin constraint and we set $\delta=1.0$ in our experiments. 

Similarly, let us define the $i$-th training image's $j$-th object-level features of $S$, $A$ and $T$ as $f_{i,j}^S$, $f_{i,j}^{A}$, and $f_{i,j}^{T}$ respectively. As the  domain-level metric regularization on object features, we will also minimize the following object-level triplet loss:    

\begin{equation}\label{equ:reg_loss_obj}
  L^{R}_{obj} = {\rm max}(d(f_{i,j}^{S}, f_{i,j}^{T} ) - d(f_{i,j}^{S}, f_{i,j}^{A}) + \delta, 0).  
\end{equation}


\subsection{Loss Function}
The final training loss of the proposed network is a summation of each individual part, which can be written as:

\begin{equation}\label{equ:all_loss}
 L = L_{cls} + L_{reg}+ w*(L_{img} + L_{obj} + L^{R}_{img} + L^{R}_{obj}), 
\end{equation}
where $L_{cls}$ and $L_{reg}$ are the loss of classification and the loss of regression in the original Faster R-CNN respectively, and  $w$ is a weight to balance the Faster R-CNN loss and the   domain adaptation loss for training. We set $w=0.1$ in our experiments. In the training, the proposed domain adaptive object detection framework  can be trained in an end-to-end manner using a standard Stochastic Gradient Descent algorithm. During the testing, the original Faster R-CNN architecture with trained adapted weights can be used for object detection, after removing the domain adaptation components.

\subsection{General Domain Adaptive Object Detection}
Our model has the capability to be adapted to general domain adaptive object detection. For the scenarios that the images from target domain are synthesized from the source domain with pixel-to-pixel correspondence (\textit{e.g.}, Cityscapes$\longrightarrow$Foggy Cityscapes), our method can be directly applied without modification. For the scenarios where target and source domains do not have strict correspondence (\textit{e.g.}, Cityscapes$\longrightarrow$KITTI), our method can be applied by simply removing the  $L^R_{obj}$ loss to eliminate the dependence on the object  alignment in the model training.

\section{Experiments}~\label{Sec:Experiments}
\vspace{-2em}
\subsection{Benchmark}
Our experiments are based on the public object detection benchmarks Cityscapes~\cite{cordts2016cityscapes} and Foggy Cityscapes~\cite{sakaridis2018semantic} for autonomous driving. Cityscapes~\cite{cordts2016cityscapes} is a widely used autonomous driving dataset, which is a collection of images with city street scenarios in clear weather conditions from 27 cities. In Cityscapes dataset, there are 2,975 training images and 500 validation images with instance segmentation annotations which can be transformed into bounding-box annotations with $8$ categories. All images are 3-channel RGB images and captured by a  car-mounted video camera with the same resolution of $1024 \times 2048$. Foggy Cityscapes~\cite{sakaridis2018semantic} is established by simulating the fog of different intensity levels on the Cityscapes images, which generates the simulated three levels of fog based on the depth map and a physical model~\cite{sakaridis2018semantic}. Its image number, resolution, training/validation split, and annotations are same as those of Cityscapes. Following the previous  methods~\cite{chen2018domain,xu2020cross,guan2021uncertainty}, the images with the fog of highest intensity level are utilized as the target domain for transfer learning in our experiments.  

\subsection{Experimental Setting}

\begin{table*}[t!]
	\caption{AP for each class and overall mAP with comparison  methods on the Cityscapes$\longrightarrow$Foggy Cityscapes experiment (\%) as clear to foggy adaptation. Note that the best performance is bold and the second best is underlined.}
	\label{tab:comparison_result}
    \small
	\centering
        \begin{tabular}{c|cccccccc|c}
        \hline \hline
        Methods                          & bus           & bicycle       & car           & mcycle        & person        & rider         & train         & truck         & mAP           \\ \hline
        SCDA-CVPR'19~\cite{zhu2019adapting}            & 39.0          & 33.6          & 48.5          & 28.0            & 33.5          & 38.0          & 23.3          & 26.5          & 33.8          \\
        DM-CVPR'19~\cite{kim2019diversify}             & 38.4          & 32.2          & 44.3          & 28.4          & 30.8          & 40.5          & 34.5          & 27.2          & 34.6          \\
        MAF-ICCV'19~\cite{he2019multi}                 & 39.9          & 33.9          & 43.9          & 29.2          & 28.2          & 39.5          & 33.3          & 23.8          & 34.0          \\
        MCAR-ECCV'20~\cite{zhao2020adaptive}            & 44.1          & 36.6          & 43.9          & \textbf{37.4}          & 32.0          & 42.1          & 43.4          & \textbf{31.3}          & 38.8          \\
        SWDA-CVPR'19~\cite{saito2019strong}             & 36.2          & 35.3          & 43.5          & 30.0          & 29.9          & 42.3          & 32.6          & 24.5          & 34.3          \\
        PDA-WACV'20~\cite{hsu2020progressive}         & 44.1          & 35.9          & \textbf{54.4} & 29.1          & 36.0         & 45.5          & 25.8          & 24.3          & 36.9          \\
        
        MTOR-CVPR-19~\cite{cai2019exploring}       & 38.6          & 35.6          & 44.0          & 28.3          & 30.6          & 41.4          & 40.6         & 21.9          & 35.1          \\
        DA-Faster-CVPR'18~\cite{chen2018domain}     & \underline{49.8}          & \underline{39.0}          & 53.0          & 28.9          & 35.7          & 45.2          & \underline{45.4}          & \underline{30.9} & 41.0          \\
        GPA-CVPR'20~\cite{xu2020cross}         & 45.7          & 38.7          & 54.1          & \underline{32.4} & 32.9          & \textbf{46.7}          & 41.1          & 24.7          & 39.5          \\
        RPN-PR-CVPR'21~\cite{zhang2021rpn}            & 43.6          & 36.8          & 50.5          & 29.7          & 33.3          &45.6          & 42.0          & 30.4         & 39.0          \\
        UaDAN-TMM'21~\cite{guan2021uncertainty}     & 49.4          & 38.9          & 53.6          &32.3          & \textbf{36.5}          & \underline{46.1}          & 42.7          & 28.9          & \underline{41.1}          \\
        \hline
        Ours  w/o Auxiliary Domain                       &48.4 &36.7 &53.5          &26.1         &\underline{36.1}    &45.9     &39.1 &29.3          &40.2
        \\
        Ours                             & \textbf{51.2} & \textbf{39.1} & \underline{54.3}          & 31.6          & \textbf{36.5} & \textbf{46.7} & \textbf{48.7} & 30.3          & \textbf{42.3} \\
        \hline
        Oracle                           & 49.9 &45.8 & 65.2          & 39.6          & 46.5    &51.3 &34.2 & 32.6          & 45.6 \\ \hline \hline
        \end{tabular}
\end{table*}

\noindent \textbf{Dataset setting}: We set the labeled training set of Cityscapes~\cite{cordts2016cityscapes} as source domain and the unlabeled training set of Foggy Cityscapes~\cite{sakaridis2018semantic} as target domain during the training. Then, the trained model is tested on the validation set of Foggy Cityscapes to report the evaluation result. We denote this  setting as the Cityscapes$\longrightarrow$Foggy Cityscapes experiment in this paper. 

\noindent \textbf{Training and parameter setting}: In the experiments, we adopt ResNet-50 as the backbone for the Faster R-CNN~\cite{ren2015faster} detection network, which is pre-trained on ImageNet. During training, following setting in~\cite{chen2018domain,ren2015faster}, the back-propagation and stochastic gradient descent (SGD) are used to optimize all the networks. The whole network is trained with an initial learning rate $0.01$ for $50k$ iterations and then reduced to $0.001$ for another $20k$ iterations. For all experiments, a weight decay of $0.0005$ and a momentum of $0.9$ are used, and each batch includes an image of source domain, an image of target domain and an image of auxiliary domain. For comparison, the $\lambda$ in the original GRL (Eq.~(\ref{equ:grl_2})) is set as 1. The hardness threshold $\alpha$ in the AdvGRL  (Eq.~(\ref{equ:ad_grl})) is set as $0.63$ by averaging the values of Eq.~(\ref{equ:img_loss}) when $P_i=0.7, G_i=1$ and $P_i=0.3, G_i=0$. Our code is implemented with PyTorch and Mask R-CNN Benchmark Toolbox~\cite{massa2018mrcnn}, and all models are trained using a GeForce RTX3090 GPU card with 24GB memory. 

\noindent \textbf{Evaluation metrics and comparison methods}: We set the Intersection over Union (IoU) threshold as $0.5$ to compute the Average Precision (AP) of each category and mean Average Precision (mAP) of all categories.
Then we compare our proposed method with some  recent domain adaptation comparison methods in our experiments, such as SCDA~\cite{zhu2019adapting}, DM~\cite{kim2019diversify}, MAF~\cite{he2019multi}, MCAR~\cite{zhao2020adaptive}, SWDA~\cite{saito2019strong}, PDA~\cite{hsu2020progressive}, RPN-PR~\cite{zhang2021rpn}, MTOR~\cite{cai2019exploring}, DA-Faster~\cite{chen2018domain}, GPA~\cite{xu2020cross}, and UaDAN~\cite{guan2021uncertainty}.

\subsection{Clear to Foggy Adaptation}

The results of weather adaptation from clear weather to foggy weather are represented in Table~\ref{tab:comparison_result}. Compared with other domain adaptation methods, we can see that our proposed method achieves the best detection performance with a mAP of $42.3\%$, which is higher than the second best method UaDAN~\cite{guan2021uncertainty} with a mAP improvement of $1.2\%$. For each category,  we can see that the proposed method is able to alleviate the domain gap over most of the categories in Foggy Cityscapes, \textit{e.g.}, bus got $51.2\%$, bicycle got $39.1\%$, person got $36.5\%$, rider got $46.7\%$, and train got $48.7\%$ as the best performance in AP, which is highlight in Table~\ref{tab:comparison_result}. The proposed method can reach the $48.7\%$ AP for the train detection in Foggy Cityscapes, compared to the $45.4\%$ AP by the second best method DA-Faster, where the proposed method is $3.3\%$ better than DA-Faster. While PDA got $54.4\%$ in car, GPA got $32.4\%$ in motorcycle, DA-faster got $30.9\%$ in truck as the best performance in some categories, the proposed method is comparable across these three categories with a minor difference. Obviously, compared to these recent domain adaptation methods, the proposed method achieves the best performance in overall mAP performance and more than half categories of Foggy Cityscapes.


\begin{table}[htb]
	\caption{Ablation study for mAP on the Cityscapes$\longrightarrow$Foggy Cityscapes experiment.}
	\label{tab:Ablation}
	\footnotesize
	\centering
        \begin{tabular}{c|ccccc}
        \hline \hline
         & img & obj & AdvGRL & \multicolumn{1}{c|}{Reg}                   & mAP   \\ \hline
        \multicolumn{1}{c|}{Source only}        &     &     &        & \multicolumn{1}{c|}{}  & 23.41 \\ \hline
        \multicolumn{1}{c|}{img+GRL}            & \checkmark   &     &        & \multicolumn{1}{c|}{}  & 38.10 \\ \hline
        \multicolumn{1}{c|}{obj+GRL}            &     & \checkmark   &        & \multicolumn{1}{c|}{}  & 38.02 \\ \hline
        \multicolumn{1}{c|}{img+obj+GRL (Baseline)}            & \checkmark  & \checkmark  &        & \multicolumn{1}{c|}{}  & 38.43 \\ \hline \hline
        \multicolumn{1}{c|}{img+obj+AdvGRL}     & \checkmark   & \checkmark   & \checkmark   & \multicolumn{1}{c|}{}  & 40.23 \\ \hline
        \multicolumn{1}{c|}{img+obj+GRL+Reg}    & \checkmark  & \checkmark   &        & \multicolumn{1}{c|}{\checkmark} & 41.97 \\ \hline
        \multicolumn{1}{c|}{img+obj+AdvGRL+Reg} & \checkmark   & \checkmark  & \checkmark   & \multicolumn{1}{c|}{\checkmark} & \textbf{42.34} \\ \hline \hline
        \end{tabular}
\end{table}

\subsection{Cross-Camera Adaptation}
To fully evaluate the proposed method, we conduct an experiment to perform the cross-camera adaptation between real-world  autonomous driving datasets with  different camera settings. To apply our method to the unaligned datasets in the  real-world, we simply remove  $L^R_{obj}$ (Eq.~\ref{equ:reg_loss_obj}) to apply our method from Cityscapes (source) to KITTI~\cite{geiger2012we} (target) datasets for cross-camera adaptation. Same as~\cite{chen2018domain}, we use KITTI training set (7,481 images of resolution $1250 \times 375$) as target domain in both adaptation and evaluation, and AP of Car on target domain is evaluated. The result is in Table~\ref{tab:comparison_result1}, where the proposed method achieved outstanding performance compared with recent SOTA methods.


\begin{table*}[t!]
    \caption{AP of Car on the Cityscapes$\longrightarrow$KITTI experiment as cross-camera adaptation.}
    \label{tab:comparison_result1}
    \footnotesize
    \centering 
        \begin{tabular}{ccccccccc}
        \hline
        \hline
        & MAF-ICCV’19\cite{he2019multi}  & ATF-ECCV'20\cite{he2020domain}  & ART-CVPR'20\cite{zheng2020cross}  &GPA-CVPR'20\cite{xu2020cross}  & SGA-TMM'21\cite{zhang2021self} &  UIT-ESwA'22\cite{arruda2022cross}        & Ours  \\ \hline
         AP     & 72.10          & 73.50         & 73.60        & 65.36       & 72.02    & 73.70       & \textbf{74.38} \\ \hline \hline
        \end{tabular} 
\end{table*}


\subsection{Ablation Study on Components}
The effect of each individual proposed component for the domain adaptation detection method is investigated in this section. All experiments are conducted with the same RestNet-50 backbone on the Cityscapes$\longrightarrow$Foggy Cityscapes experiment. The results are presented in Table~\ref{tab:Ablation}. In the first row, `img' and `obj' stand for the image-level adaptation module and object-level adaptation module respectively, while `AdvGRL' and `Reg' denote the proposed Adversarial Gradient Reversal Layer and domain-level metric Regularization respectively. `img+obj+GRL' stands for the baseline model in our experiment. We denote that `img+obj+AdvGRL' (Ours  w/o Auxiliary Domain) and `img+obj +AdvGRL+Reg' use the AdvGRL to replace the original GRL. The `Source only' indicates the Faster R-CNN model without domain adaptation trained only with labeled source domain images. The ablation study in Table~\ref{tab:Ablation}  clearly  justifies the positive effect of each proposed component of the domain adaptive object detection. 



\subsection{Ablation Study on Parameters}
The study on different hyper-parameters of Eq.~\ref{equ:all_loss} and Eq.~\ref{equ:ad_grl} are   conducted. We use the Cityscapes$\longrightarrow$Foggy Cityscapes as the study case. First, the loss balance weight $w$ in Eq.~\ref{equ:all_loss} is set as $0.1$, $0.01$, $0.001$ separately for training, and the corresponding detection mAP(s) are 42.34, 41.30, 41.19, respectively. Second, in the AdvGRL (Eq.~\ref{equ:ad_grl}), the overflow threshold $\beta$ and hardness threshold $\alpha$ are set as (1) $\beta=30, \alpha=0.63$, (2) $\beta=10,  \alpha=0.63$, (3) $\beta=30, \alpha=0.54$, and (4) $\beta=10, \alpha=0.54$, where $\alpha=0.54$ is computed by averaging the values of  Eq.~\ref{equ:img_loss} when $P_i=0.9, G_i=1$ and $P_i=0.1, G_i=0$. The detection mAP(s) of these settings are (1) 42.34, (2) 38.83, (3) 39.38, (4) 40.47, respectively.

\subsection{Discussion on Visualized Hard Examples}
\begin{figure}[t]
  \centering
  \includegraphics[width=0.95\linewidth]{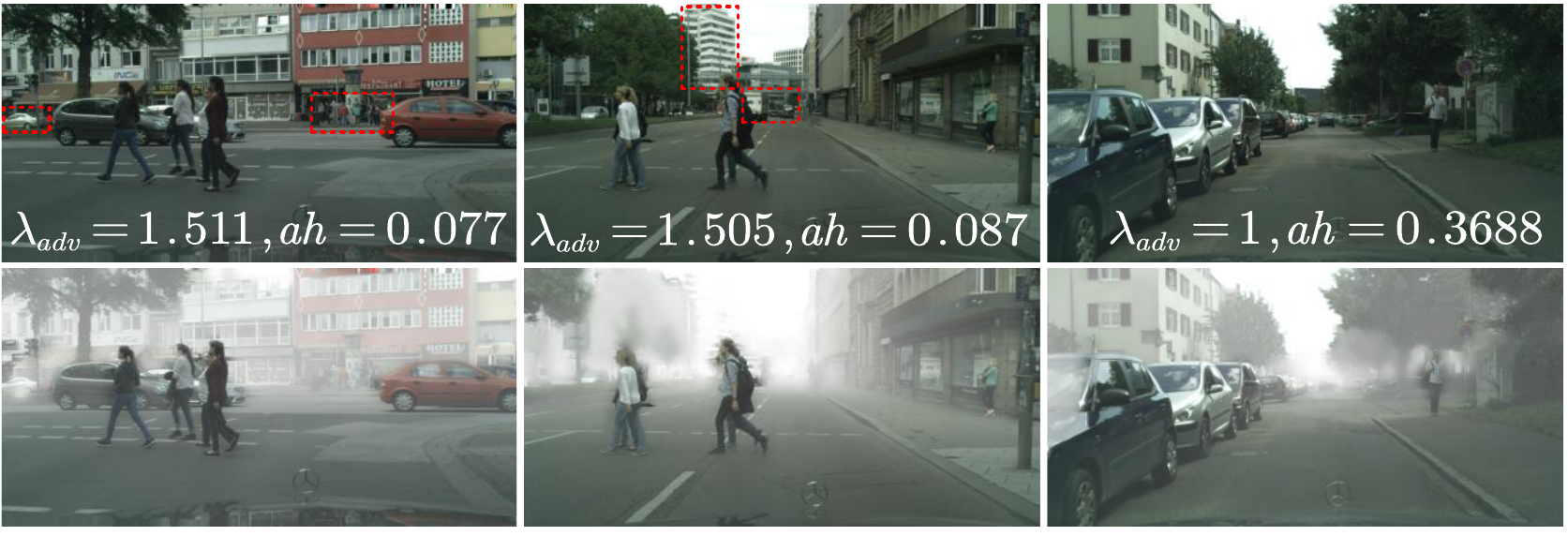}
 \caption{Hard examples (bigger  $\lambda_{adv}$) mined by AdvGRL. Left to right: two mined hard examples, an    easy example.} 
  \label{fig:hard-examples}
\end{figure}

Using $\lambda_{adv}$ of the proposed AdvGRL, we could find some hard examples, as shown in Fig.~\ref{fig:hard-examples}. We compute the $L_1$ distance of features $F^S_i$ and $F^T_i$ after the CNN backbone of  Fig.~\ref{fig:da_faster_rcnn} as the approximated hardness $ah$, where smaller $ah$ means harder for transfer learning. Intuitively, if the fog covers more objects as shown in bounding-box regions of  Fig.~\ref{fig:hard-examples}, it will be more difficult. 

\subsection{Discussion on Domain Randomization,  Pre-trained Models, and Qualitative Results}
\noindent \textbf{Domain Randomization:} Domain randomization might be used to reduce the domain shift between source domain and target domain. We use two ways as the domain randomization in the Cityscapes$\longrightarrow$Foggy Cityscapes experiment,  \textit{i.e.}, regular data argumentation and  CycleGAN~\cite{zhu2017unpaired} based image style transfer. 1) We construct the auxiliary domain by regular data argumentation (color change + blur + salt \& pepper noises), where our method's detection mAP is 38.7, compared to our 42.3 by the  auxiliary domain with rain synthesis. 2) We train a CycleGAN to transfer the image style between the training sets of Cityscapes and Foggy Cityscapes. Using the generated fake foggy-style image of Cityscapes by the trained CycleGAN model to train a Faster R-CNN model, it achieves detection mAP as 32.8. These experiments show that commonly used domain randomization could not well solve the domain adaptation problem.   

\noindent \textbf{Pre-trained Models:} We use the pre-trained Faster R-CNN model in~\cite{chen2018domain} to initialize our method, then our method gets the detection mAP as 41.3 in the Cityscapes$\longrightarrow$Foggy Cityscapes experiment, compared to 42.3 by our method without pre-trained detection model. 

\begin{figure}[htb]
	\centering
	\includegraphics[width=0.95\columnwidth]{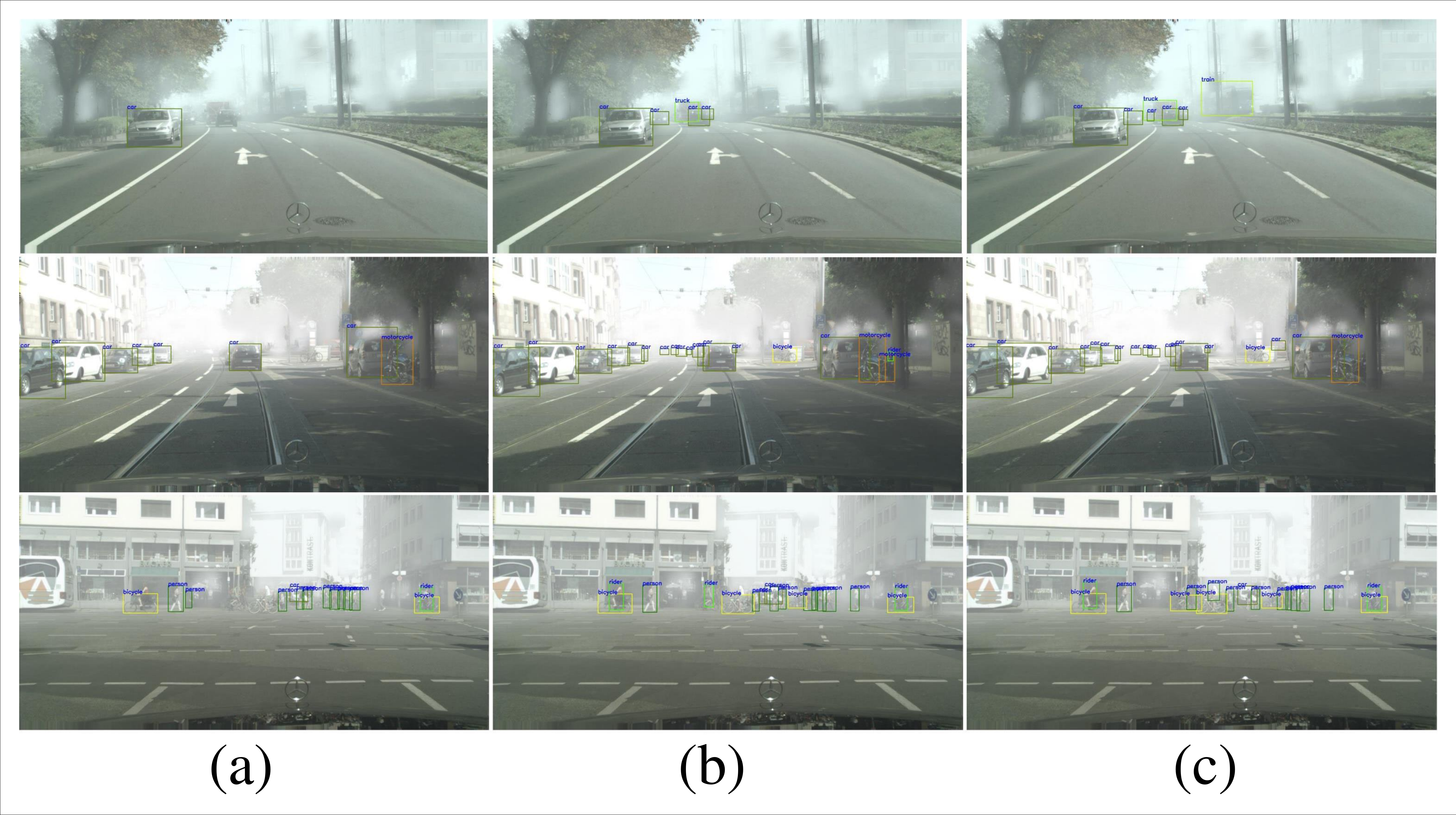}
	\caption{Qualitative visualization results on validation set of Foggy Cityscapes: (a) Original Faster R-CNN without domain adaptation, (b) Faster R-CNN with image-level and object-level adaptations using GRL (Baseline), (c) Proposed Method. Note: different colors represent different categories.} 
	\label{fig:visual_detection} 
\end{figure}

\noindent \textbf{Qualitative Results:} We visualize some detection results on the Foggy Cityscapes dataset in  Fig.~\ref{fig:visual_detection}, which shows that the proposed domain adaptive method improves the detection  performance in foggy weather significantly.   

\vspace{-1.5em}
\section{Conclusions}\label{Sec:Conclusions}

In this paper, we propose a novel domain adaptive object detection framework for autonomous driving. The image-level and object-level adaptations are designed to reduce the domain shift on the global image style and local object appearance. A new adversarial gradient reversal layer is proposed to perform adversarial mining for hard examples together with domain adaptation. Considering the feature metric distance between the source domain, target domain, and auxiliary domain by data augmentation, we propose a new domain-level metric regularization. Furthermore, our method could be applied to solve the general domain adaptive object detection problem. We conduct the transfer learning experiments from Cityscapes to Foggy Cityscapes and from Cityscapes to KITTI, and experimental results show that the proposed method is quite effective. 

\noindent \textbf{Acknowledgement:} This work was supported by NSF 2215388.

{\small
\bibliographystyle{ieee_fullname}
\bibliography{egbib}
}

\end{document}